# Deep Learning Prediction of Severe Health Risks for Pediatric COVID-19 Patients with a Large Feature Set in 2021 BARDA Data Challenge

Sajid Mahmud[&], Elham Soltanikazemi[&], Frimpong Boadu[&], Ashwin Dhakal[&], Jianlin Cheng*

Department of Electrical Engineering and Computer Science, University of Missouri, Columbia, MO 65211, USA

&: joint first author
*: corresponding author (chengji@missouri.edu)

## Abstract

Most children infected with COVID-19 have no or mild symptoms and can recover automatically by themselves, but some pediatric COVID-19 patients need to be hospitalized or even to receive intensive medical care (e.g., invasive mechanical ventilation or cardiovascular support) to recover from the illnesses. Therefore, it is critical to predict the severe health risk that COVID-19 infection poses to children to provide precise and timely medical care for vulnerable pediatric COVID-19 patients. However, predicting the severe health risk for COVID-19 patients including children remains a significant challenge because many underlying medical factors affecting the risk are still largely unknown. In this work, instead of searching for a small number of most useful features to make prediction, we design a novel large-scale bag-of-words like method to represent various medical conditions and measurements of COVID-19 patients. After some simple feature filtering based on logistical regression, the large set of features is used with a deep learning method to predict both the hospitalization risk for COVID-19 infected children and the severe complication risk for the hospitalized pediatric COVID-19 patients. The method was trained and tested the datasets of the Biomedical Advanced Research and Development Authority (BARDA) Pediatric COVID-19 Data Challenge held from Sept. 15 to Dec. 17, 2021. The results show that the approach can rather accurately predict the risk of hospitalization and severe complication for pediatric COVID-19 patients and deep learning is more accurate than other machine learning methods.

## Introduction

Predicting the severe health risks for pediatric COVID-19 (coronavirus disease 2019) patients is important for providing timely medical care to reduce severe outcomes. However, many medical and health conditions affecting the risks are still largely unknown and accurate prediction of the health risk of COVID-19 infection remains a significant challenge [1], [2]. To stimulate the development of methods to predict severe medical risks for pediatric COVID-19 patients [3], [4], in 2021 the Biomedical Advanced Research and Development Authority (BARDA) organized a data challenge for the community to develop predictive methods to tackle the problem and test them on the large amount of medical data of the pediatric COVID-19 patients provided by the National COVID Cohort Collaborative (N3C) [5].

COVID-19 patients have different kinds of medical conditions and measurements stored in various medical data tables [6]. The patients as a whole may have tens of thousands of medical conditions and measurements, but one patient usually only has a small number of conditions and measurements. The number, types and time stamps of conditions and measurements of individual patients also vary a lot. All these make it difficult to apply machine learning methods to predict the risks of the disease because machine learning usually requires that the features describing individual patients can be compared. Therefore, a key challenge is to create a uniform representation of the features for all the patients (e.g., a standardized feature vector representing medical conditions and measurements) that can be used by machine learning methods to integrate the patient information to predict the severe medical risk (i.e., hospitalization risk or life-threatening complication risk). Another key challenge is that it is not clear what medical features (or factors) out of thousands of conditions or measurements are relevant to the prediction [7].

To create a uniform representation of medical conditions and measurements of patients without missing important features, we use all the unique concept ids defined by the OMOP common data model [8] as well as their corresponding concept set ids describing the conditions and measurements of all the patients in the training data provided by the BARDA Data Challenge as the initial set of features to represent their medical information. This approach is similar to the "bag-of-words" method [9] widely used to represent a document in text classification. The concept ids for patients are like words for documents. The frequency of a concept id or concept set id (i.e., feature) of six kinds of medical data (e.g., condition, observation, procedure, drug, device, and visit) is used as the value of the feature. The value of the latest measurement of each unique measurement concept id is used as the value of the measurement feature. To facilitate machine learning, the value of a feature is normalized as a non-zero number between 0 and 1 by the minimum and maximum values of the feature in the training data, while 0 is reserved to represent a missing value.

The number of the initial features is very large because there are many unique concept ids and concept set ids. To filter out some irrelevant features, a logistic regression method is used to assess the value of each type of features for predicting the outcome to decide which kind of features is included into the final set. Based on the logistic regression prediction results, some features that have little predictive power are removed.

The final features of all the patients are used to train and validate several machine learning methods to predict their hospitalization risk. Among all the methods tested (i.e., logistic regression [10], support vector machines (SVM) [11], random forest [12], and deep learning [13]), the deep learning model works best. The deep learning method with the careful training and validation can handle the extremely unbalanced data (e.g., only ~2.7% positive examples in the hospitalization risk prediction dataset or ~8.8% of the positive examples in the severe complication risk prediction dataset) better than the other methods.

## Methods

### Datasets

The BARDA COVID-19 Data Challenge provides the datasets for two prediction tasks. The first task is to predict if a child infected by COVID-19 will be hospitalized. Various medical data stored in OMOP tables such as condition, procedure, measurement, observation, drug, devices, payer, visit, health providers, medical sites, and personal information (age, race, gender) for 119,151 patients are provided. The patients are labeled into two categories: hospitalized or not hospitalized.

The second task is to predict if a COVID-19 hospitalized child will have life-threatening complication and needs life-support medical intervention (e.g., invasive mechanical ventilation or cardiovascular support). The same kind of medical data in Task 1 is provided for 9,545 hospitalized patients of Task 2. The patients are labeled into two categories: severe complication or no severe complication. The overall pipeline of the process is shown in **Figure 1**.

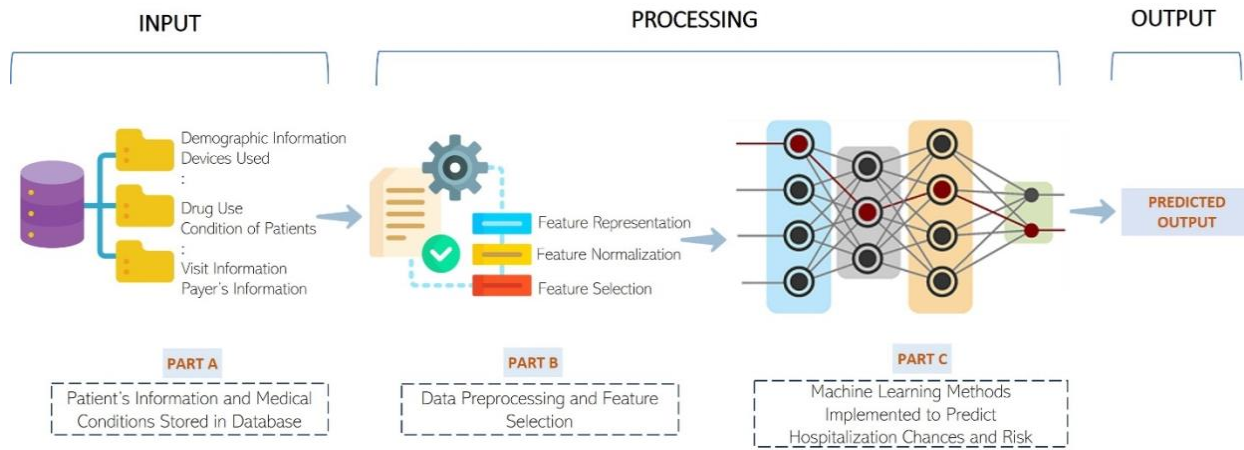

**Figure 1**: A high level overview of the end-to-end hospitalization and risk prediction pipeline.

**A large-scale uniform representation of patient data and feature normalization**

For Task 1, nine kinds of medical data (i.e., condition, procedure, measurement, observation, drug, devices, payer, visit, and person information (age, race, gender)) (as shown in PART A, Figure 1) are used to generate initial features to represent each patient. Some data such as care site and provider are not considered because we aim to develop a general method to predict the hospitalization risk regardless of the external factors such as medical sites and providers. The unique concept ids of each kind of data (e.g., condition) in the training data are used as distinct features to describe the data. For instance, 15,403 unique condition concept ids occurring in all the patients in the training data are used as 15,403 features to represent the medical conditions. Moreover, the concept ids of condition, procedure, observation, drug, and devices are mapped to the higher-level concept set ids, leveraging the known categorizations of the concept ids into broader categories. The unique concept set ids of the five kinds of data (condition, procedure, observation, drug, and device) are also used as the features. For example, 1,070 unique condition concept set ids are used as features. The concept set ids of measurements are not used because a measurement value is usually only relevant for a specific measurement type. The frequency (count) of a discrete feature (e.g., a condition concept id or concept set id) for each patient is used as its feature value. The continuous value of a measurement is used as its value. The frequency of the concept ids for the demography features (i.e., race, ethnicity, and gender) is always binary, i.e., only one of the concept id for each type of demography information (e.g., gender) is 1 for each person, resulting in one-hot representation for each kind of demographic information. Finally, the age of a person is also used as an initial feature. The representation above results in a very large feature set consisting of 41,780 features for 119,151 patients in total.

The feature vector of each patient is very sparse, i.e., only a small number of features have value, while most features do not have a value. Therefore, we do not try to impute the missing values. Instead, we use 0 to represent all the missing values and use the normalized positive values to represent the observed values of the features. The discrete features (e.g., condition, procedure, drug, observation, device, and visit) is normalized by a customized min-max normalization formula (normalized_value = (initial_value - min + 1) / (max-min + 1)), where min and max denote the minimum and maximum value of the feature in the training data. 1 is added here to make sure every value including min has a non-zero positive normalized value in the range (0, 1] because 0 is reserved for the missing value. The value of measurement features is normalized by a customized min-max formula: normalized_value = (initial_value - min + c) / (max-min +c), where c = (max-min) / 10. Like the normalization of discrete features, c is added here to make sure each measure value has a non-zero positive normalized value. The demographic features do not need to be normalized because one-hot encoding is applied to them. The age is normalized by diving it by 19. After the normalization, all the features have a value between 0 and 1, which is ideal for function-approximation machine learning methods such as logistic regression and deep learning. The feature normalization strategies for different features are portraited in **Figure 2**.

The data representation used in Task 2 is similar to Task 1 with some minor conceptual differences. Because the training datasets used for Task 1 and Task 2 are different, the actual number of features (concept ids or concept set ids) extracted from the training data are different. Age is not used in Task 2 as it does not add value on top of other features according to the experiments. The feature set of Task 2 consists of 31,590 features in total. As shown in **Figure 2**, the same normalization method used in Task 1 is applied to Task 2.

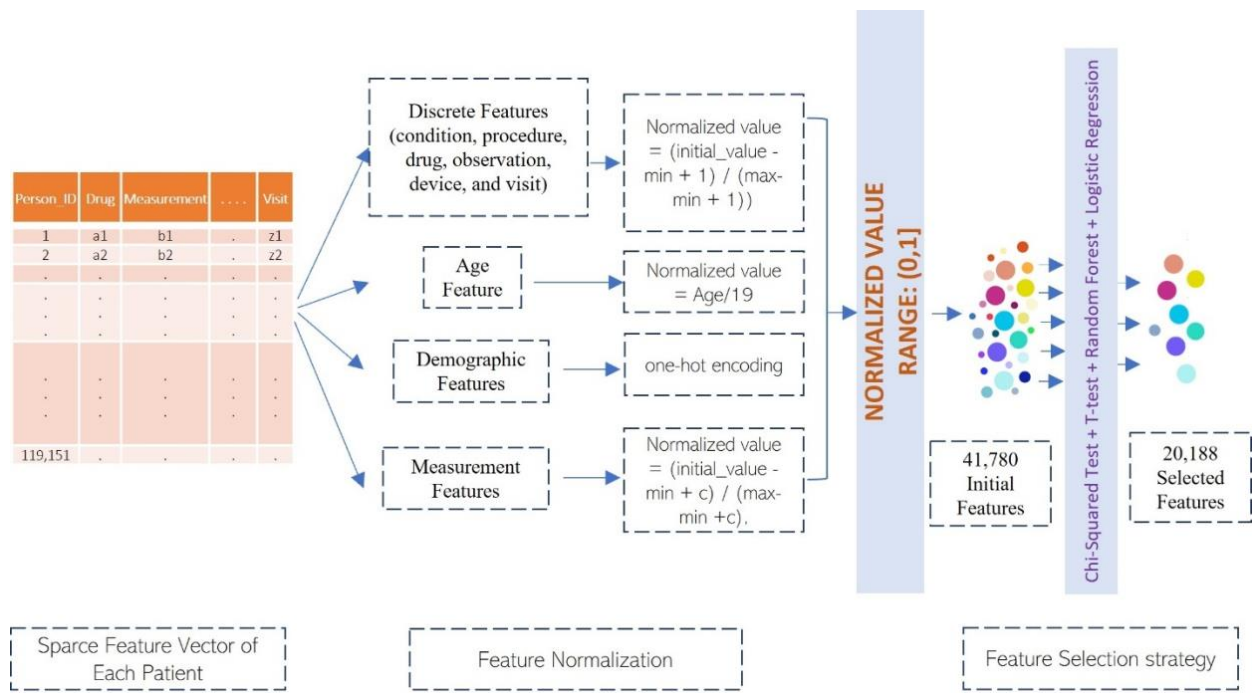

**Figure 2**: The pipeline of data preprocessing and feature selection.

**Feature selection**

A large feature set containing irrelevant features can mislead training and reduce the accuracy of machine learning methods. Moreover, too many features require a lot of memory to store them and slow down machine learning training [14]–[16]. We use a fast, easy-to-implement logistic regression method to assess the predictive capability of each kind of feature for the prediction task. For the same type of features (e.g., condition concept id and condition concept set id), the logistic regression is applied to them separately and their combination to evaluate if they are redundant to each other.

For Task 1, condition concept id, measurement concept id, observation concept id, observation concept set id, procedure concept id, drug concept id, drug concept set id, device concept id, visit concept id have some predictive value and are kept in the final feature set. The demography features (e.g., race, gender, ethnicity), payer features, and age feature have relatively lower predictive value and are excluded. Some concept set ids such as condition concept set id have some predictive value, but do not add value on top of their corresponding concept id and therefore are excluded. As the concept ids occurring in positive patients are likely more useful to recall the positive examples, the concept ids that only occur in the negative examples are excluded. After the feature selection above, 20,188 features are left in the final feature set, including 7250 condition features, 194 device features, 1396 observation features, 194 observation set features, 2834 procedure features, 4546 drug features, 579 drug set features, 3167 measurement features, and 28 visit features. In addition to the approach above, we also use the Chi-Square test, t-test, random forest, and weights of logistic regression to select features. But due to limited experimentation, they do not generate a smaller set of features obviously better than 20,188 features above. The 20,188 features are used to train and validate the machine learning methods in this work.

The same feature selection method used in Task 1 is applied to Task 2. Condition concept id, measurement concept id, observation concept id, observation concept set id, procedure concept id, drug concept id, drug concept set id, device concept id, visit concept id have some predictive value and are kept in the final feature set. The demography features are also included. Some concept set ids such as drug concept set id have some predictive value, but do not add value on top of drug concept id and therefore are excluded. As the concept ids occurring in positive patients are likely more useful to recall the positive examples, all of them are used as features. Concept ids that only occur in negative examples are also assessed. The concept ids of some data types only occur once in the negative examples are excluded, while some concept ids that occur at least twice in only negative examples are kept. After the feature selection above, 24,807 features are left in the final feature set, including 8571 condition features, 1449 observation features, 5422 procedure features, 4979 drug features, 4328 measurement features, 33 visit features, and 25 demography features (race and gender). The ethnicity information is not used because it is very incomplete and is largely redundant to the race feature.

**Deep learning method**

We test different machine learning methods (e.g., logistic regression, ridge regression, Lasso, random forest, SVM, and deep learning) on the features of Task 1 or Task 2. The deep learning method works best and can deal with the extremely imbalanced data (e.g., only ~2.7% of examples are positive in Task 1) better. Therefore, we focus on describing the deep learning method below.

We design and test different deep learning architectures and eventually settle down on a network (**Figure 3**) that has one input layer, one dense hidden layer with 100 hidden nodes and Relu activation function [17], a batch normalization layer, one dropout layer (drop out rate: 0.5), one hidden layer with 100 hidden nodes and Relu function, one dropout layer (drop out rate: 0.5), one hidden layer with 64 hidden nodes, Relu function and weight regularization, one dropout layer (drop out rate: 0.5), and an output layer with sigmoid function [18] to predict the hospitalization risk for Task 1 or severe complication/outcome risk for Task 2. The overall neural network architecture is shown in Figure 3. The deep model was implemented on top of the TensorFlow deep learning library. The network is trained by ADAM optimization. The initial learning rate is set to 0.0001. It is worth noting that the hyper-parameter and the number of layers of this network have not been extensively tuned due to the time and computing resource limit.

For Task 1, a major challenge of training a machine learning method on this data is how to handle the extremely low proportion of positive examples (i.e., ~ 2.7%). We try weighting positive class more, oversampling positive examples and undersampling negative examples. The undersampling approach appears to work best under the constraint of the limited time and computer memory [2]. In this approach, we set 10% of data as validation data and 90% as training data. We further randomly sample 10% of negative examples in the training data and combine them with all the positive examples in the training data as the final training data to train the deep learning model. Early stopping monitoring the area under the precision-recall curve (auPRC) on the validation data is used to control when to stop training and which deep model is selected. If there is no increase of auPRC over 20 consecutive epochs, the training stops and the model with the highest auPRC score during the training is kept. The training and validation procedure based on the same validation data above is repeated 10 times to produce 10 deep models. The model with the highest auPRC score is selected as the final model for a round of training and validation (called a trial). We repeat the training and validation trial above 10 times by randomly splitting training and validation data to cover a larger space of the training data. This process produces 10 trained deep models, which are used as an ensemble to make predictions for any new test data. The average of the prediction made by the 10 deep models is used as the final prediction. In our experiment, the ensemble predictor is generally more accurate than an individual deep model by up to a few percentage points.

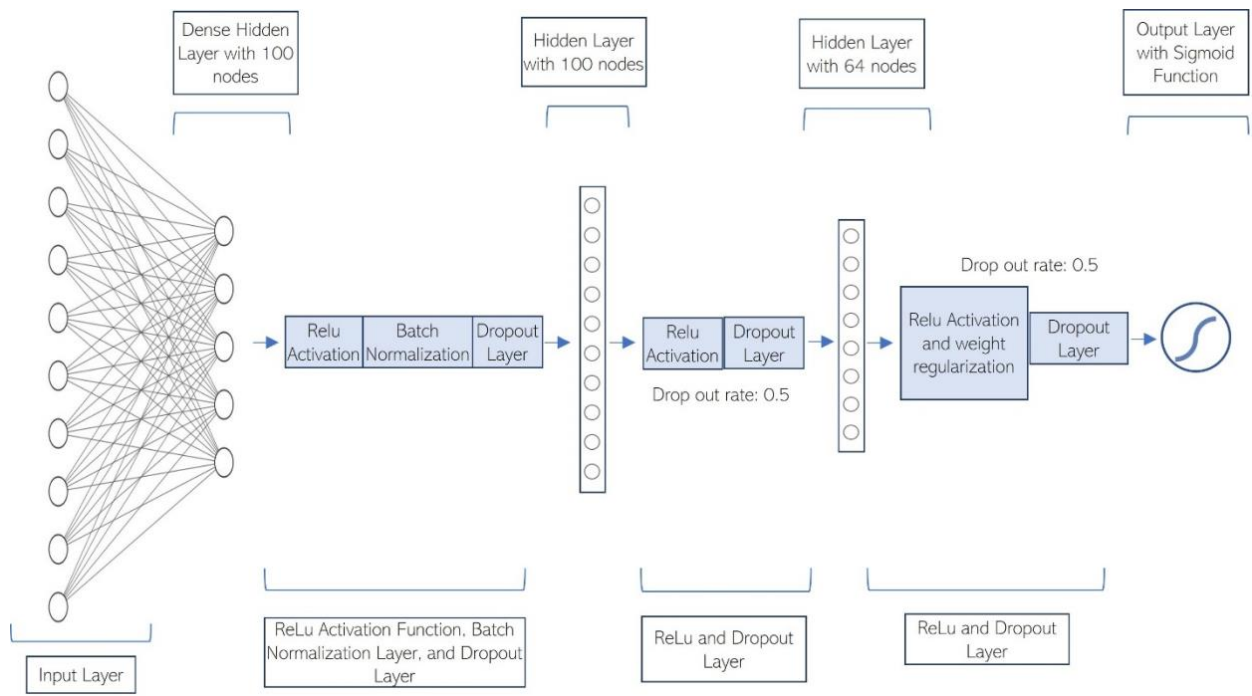

**Figure 3**: The deep neural network architecture to predict both the hospitalization risk for COVID-19 infected children and the severe complication risk for the hospitalized pediatric COVID-19 patients.

For Task 2, no under-sampling of negative examples is needed because it has a higher percentage of positive examples (~8.8%). To train and validate a deep model, we divide the data into a training dataset and a validation dataset according to 95%-5% ratio. To avoid overfitting, early stopping is applied. The auPRC score (area under the precision-recall curve) on the validation dataset is monitored. The deep model with the highest auPRC score on the validation data is kept. We repeat the training and validation trial above 25 times by randomly splitting training and validation data. The top 20 models with the highest sum of auPRC, auROC, and F2 score are stored to form an ensemble for the prediction use. The average of the predictions of the 20 models is used as the final prediction.

## Results

### Evaluation Metrics

We use three metrics to evaluate the predictions: auROC, auPRC, and F2 measure score recommended by the BARDA Data Challenge. auROC is the area under the ROC curve. auPRC is the area of the precision-recall curve. F2 measure score is 5 * precision * recall / (4 * precision + recall). F2 measure weighs recall more than precision, which is different from the standard F1 measure treating them equally.

**Table 1**. The accuracy of the deep learning method for predicting the hospitalization risk in Task 1 and the severe complication risk in Task 2.

| Task | auROC | auPRC | F2 measure |
|---|---|---|---|
| Task 1 | 0.88 | 0.49 | 0.62 |
| Task 2 | 0.90 | 0.59 | 0.63 |

### Results of Prediction of Hospitalization Risk (Task 1)

We train and validate the deep learning method on the training data of Task 1 minus the test data (about 200 test examples) provided by the BARDA Data Challenge first and then blindly tested it on the test data. The auROC, auPRC and F2 measure on the test data are 0.88, 0.49, and 0.62 (**Table 1**), respectively. The accuracy of the deep learning method is much higher than other tested machine learning methods such as the logistic regression method trained and tested on the same datasets (data not shown).

### Results of Prediction of Severe Complication Risk (Task 2)

We train and validate the deep learning method on the training data of Task 2 minus the test data (about 200 test examples) provided by the BARDA Data Challenge first and then blindly test it on the test data. The auROC, auPRC and F2 measure on the test data are 0.90, 0.59, and 0.63 (**Table 1**), respectively. The results are also much better than the logistic regression trained and tested on the same datasets (data not shown).

## Conclusion and Future Work

In this work, we design a novel uniform representation for different sources of medical data of patients without manually selecting any features based on prior knowledge, aiming to capture most information relevant to the prediction of the risk of severe outcomes posed to the pediatric COVID-19 patients. The uniform feature representation and the feature normalization make various machine learning methods applicable to this problem. However, although being rather complete, the feature set is very large and contains irrelevant features, which need to be reduced to improve the downstream prediction performance and speed up training. The logistic regression is used as a quick way to assess the predictive power of different types of features. However, more feature selection methods and experiments are needed to further improve feature selection.

Measurement, condition, procedure, drug, observation and visit features have relatively more predictive power that other features. Device features are also useful but can be only applied to a smaller portion of patients that have the device information. Moreover, our experiments show that different types of features are complementary because combining them together yields substantially better performance than using a single type of features. We believe the novel feature representation developed in this work can be applied to predict the risks of other diseases.

Moreover, our experiment demonstrates that deep learning is an effective approach to the prediction of COVID-19 risks. It generalizes better to the new data and handles the extremely unbalanced data much better than other methods. Its training is also relatively fast. Its performance is stable when training data is changed. However, more extensive search for better deep learning architectures and hyperparameters is needed to obtain better performance. A future study of the impact of different features on the prediction would be important for explaining/interpreting their medical meaning and making them useful for preventing and alleviating the severe outcome of the COVID-19 infection in clinical practice.

## Acknowledgement

The analysis and method described in this work was conducted in support from the Department of Health and Human Services, Office of the Assistant Secretary for Preparedness and Response, Biomedical Advanced Research and Development Authority (BARDA) 2021 Pediatric COVID-19 Data Challenge. We thank the BARDA Data Challenge organizers and N3C for providing the anonymized COVID-19 patient data and the computation platform used in this work. The analyses described in this publication were conducted with data or tools accessed through the NCATS N3C Data Enclave covid.cd2h.org/enclave and supported by CD2H - The National COVID Cohort Collaborative (N3C) IDeA CTR Collaboration 3U24TR002306-04S2 NCATS U24 TR002306. This research was possible because of the patients whose information is included within the data from participating organizations (covid.cd2h.org/dtas) and the organizations and scientists (covid.cd2h.org/duas) who have contributed to the on-going development of this community resource (cite this https://doi.org/10.1093/jamia/ocaa196).

## Disclaimer